\documentclass[10pt,twocolumn,letterpaper]{article}

\usepackage{cvpr}              

\usepackage[accsupp]{axessibility}  

\usepackage{multirow}

\definecolor{cvprblue}{rgb}{0.21,0.49,0.74}
\usepackage[breaklinks,colorlinks,allcolors=cvprblue]{hyperref}

\newcommand{\notsosmall}{\fontsize{10pt}{12pt}\selectfont}

\def\paperID{4544} 
\def\confName{CVPR}
\def\confYear{2026}

\title{Ov3R: Open-Vocabulary Semantic 3D Reconstruction from RGB Videos}

\author{
	Ziren Gong$^{1}$ \quad\quad\quad Xiaohan Li$^{2}$ \quad\quad\quad Fabio Tosi$^{1}$ \quad\quad\quad Jiawei Han$^{3}$ \\ Stefano Mattoccia$^{1}$ \quad\quad\quad Jianfei Cai$^{4}$ \quad\quad\quad Matteo Poggi$^{1}$ \smallskip \\
	\notsosmall \hspace{-0.3cm}$^{1}$University of Bologna \quad\quad $^{2}$The University of Hong Kong \quad\quad $^{3}$Beijing Institute of Technology \quad\quad $^{4}$Monash University \smallskip \\
    \texttt{Project page:} \url{https://zorangong.github.io/Ov3R_page/}
}

\begin{document}

\twocolumn[{
\renewcommand\twocolumn[1][]{#1}
\maketitle
\begin{center} 
    \vspace{-0.7cm}
    \centering
    \includegraphics[width=\linewidth,scale=1.00]{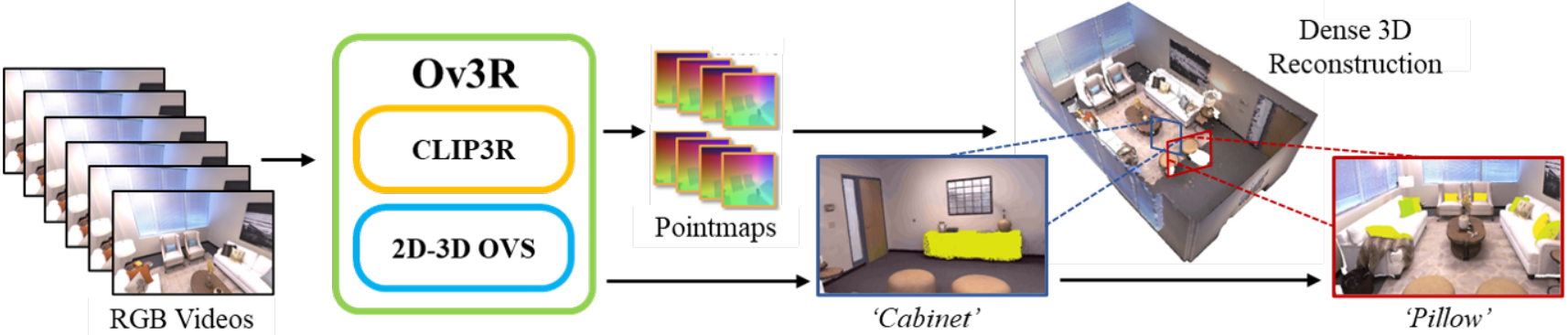}
    \vspace{-0.7cm}
    \captionof{figure}{\textbf{Ov3R is an Open-Vocabulary Semantic 3D Reconstruction Framework.} It consists of two novel feed-forward modules, \textit{CLIP3R} and \textit{2D–3D OVS}, and excels in both 3D reconstruction and open-vocabulary 3D semantic segmentation.}\vspace{0.3cm}
    \label{fig:teaser}
\end{center}
}
]

\begin{abstract}
We present Ov3R, a novel framework for open-vocabulary semantic 3D reconstruction from RGB video streams, designed to advance Spatial AI. 
The system features two key components: \textit{CLIP3R}, a CLIP-informed 3D reconstruction module that predicts dense point maps from overlapping clips alongside object-level semantics; and \textit{2D–3D OVS}, a 2D-3D open-vocabulary semantic module that lifts 2D features into 3D by learning fused descriptors integrating spatial, geometric, and semantic cues. Unlike prior methods, Ov3R incorporates CLIP semantics directly into the reconstruction process, enabling globally consistent geometry and fine-grained semantic alignment. Our framework achieves state-of-the-art performance in both dense 3D reconstruction and open-vocabulary 3D segmentation. 
\end{abstract}    
\section{Introduction}



Spatial AI systems \cite{davison2018futuremapping} aim to understand both the geometry and semantics of the surrounding environment from images in real-time, enabling an embedded AI agent to navigate and interact with it effectively.

Dense 3D reconstruction lies at the core of these systems. Over the decades, this task has taken various forms — from Structure-from-Motion (SFM) \cite{lindenberger2021pixel, liu20233d, schonberger2016structure, snavely2006photo} and Multi-View Stereo (MVS), typically performed offline, to Simultaneous Localization and Mapping (SLAM) \cite{campos2021orb, engel2014lsd, gong2025hs, teed2021droid}, which is executed online by autonomous agents and thus requires real-time processing.
The latter flavor is the most suitable approach for developing Spatial AI systems, although it poses greater challenges compared to offline methods, as input images are collected incrementally rather than being available in advance -- making the system susceptible to noise, motion blur, and cumulative drift over time.

In the last few years, SLAM has undergone a resurgence, fueled by advancements in related fields such as novel view synthesis \cite{mildenhall2021nerf,kerbl20233d} and, more recently, the emergence of 3D reconstruction models \cite{wang2024dust3r} (3R), which have redefined the classical reconstruction paradigm by bypassing explicit camera pose estimation.
The former has led to a family of SLAM methods \cite{tosi2024nerfs}, based on NeRF or 3DGS, yielding dense 3D reconstructions and accurate camera tracking, albeit with high computational cost due to the need for scene-specific training. The latter, in contrast, has resulted in the first 3R-based systems capable of real-time performance, though often at the expense of tracking accuracy, due to the absence of explicit pose estimation. 
In both directions, however, the semantic understanding of the surrounding environment has remained largely unexplored -- and where considered, it has been limited to simplified scenarios with a predefined set of classes \cite{zhu2024sni,zhu2024semgauss}. As a result, a significant gap between existing SLAM methods and envisioned Spatial AI systems still persists.




In parallel, with the emergence of CLIP \cite{radford2021learning}, research on open-vocabulary 3D semantic understanding has surged. However, existing approaches largely rely on offline reconstruction pipelines \cite{kerr2023lerf, nguyen2024open3dis, peng2023openscene, qin2024langsplat, takmaz2023openmask3d, tie20242} or RGBD SLAM methods that require depth sensors \cite{martins2024ovo}, and therefore do not address the aforementioned gap. 

In this paper, we introduce Ov3R, an open-vocabulary semantic 3D reconstruction framework that processes RGB-only video streams. Ov3R comprises two key components to achieve both accurate 3D reconstruction and semantic scene understanding: (i) CLIP3R, a CLIP-informed 3R model enriched with object-level CLIP features that encode the semantics of individual objects in the scene; and (ii) 2D-3D OVS, a module that integrates rich semantic information from various foundation models — CLIP3R itself, DINO and a distilled 3D-CLIP encoder — to effectively segment the dense 3D reconstruction produced by CLIP3R. 
These two components are loosely coupled through the CLIP features, making Ov3R a modular framework capable of performing the two tasks either jointly or independently -- e.g., running only 3D reconstruction when semantic understanding is not required or vice-versa.

Through extensive experiments on 3D reconstruction and open-vocabulary 3D segmentation, we demonstrate that Ov3R delivers high-quality dense scene reconstructions along with accurate open-vocabulary 3D semantic segmentation. Our main contributions are as follows:

\begin{itemize}
    \item We present Ov3R, a novel framework that unifies 3R models and open-vocabulary 3D semantic segmentation.
    \item We design CLIP3R, a CLIP-informed 3D reconstruction model that enforces semantic consistency across images and learns stronger semantic–geometric alignment.
    \item We propose 2D–3D OVS, a module that effectively lifts 2D CLIP descriptors into fused 2D–3D representations, enabling both spatial and geometric awareness to anchor language semantics and scene geometry. 
    \item Ov3R achieves state-of-the-art performance in  3D reconstruction and open-vocabulary 3D segmentation.

\end{itemize}
\section{Related Work}


\subsection{3D Reconstruction} 


\textbf{Offline Pipelines}. This category includes strategies that assume the availability of a complete set of images beforehand 
\cite{lindenberger2021pixel, liu20233d, schonberger2016structure, snavely2006photo}, from which to process either camera parameters and 3D points via Structure from Motion (SfM) \cite{arrigoni2025taxonomy}, or 3D points only through Multi-View Stereo (MVS) \cite{wang2024learning}. With the advent of NeRF \cite{mildenhall2021nerf} and 3DGS \cite{kerbl20233d} for novel view synthesis, several derived frameworks \cite{li2023neuralangelo, chen2024mvsplat, chen2024mvsplat360, guedon2024sugar,chen2024pgsr} have been developed to achieve high-fidelity 3D reconstructions. 
However, offline processing does not fulfill the requirements of a Spatial AI system.


\textbf{SLAM Systems}. These approaches estimate the camera pose and the 3D structure of the environment during navigation -- i.e., while images are being collected in real-time.
Early hand-crafted works \cite{campos2021orb, teed2021droid, bailey2006simultaneous, durrant2006simultaneous, engel2014lsd} 
focused mostly on camera tracking, while yielding sparse 3D reconstructions.
More recently, 
the integration of novel view synthesis techniques (NeRF and 3DGS) 
into SLAM \cite{tosi2024nerfs,deng2025best} has enabled both accurate and dense reconstructions, either assuming the availability of depth sensors \cite{zhang2023go, li2024cto, li2021po} or not \cite{zhang2023go, zhu2024nicer, li2023dense, zhu2024mgs, li2023dense,li2025stereo}. Although some of these frameworks have been enhanced with semantics understanding \cite{zhu2024sni,zhu2024semgauss,gong2025dino}, their scope is often limited to a fixed set of classes and lacks real-time capabilities.

\textbf{3R-driven Models}. DUSt3R \cite{wang2024dust3r} pioneered the first end-to-end model for 3D reconstruction from two uncalibrated RGB images, predicting two pointmaps aligned to the same reference system, from which intrinsic and extrinsic camera parameters can also be derived. 
Several extensions of DUSt3R target explicit features matching \cite{leroy2024grounding},
single-image metric depth estimation \cite{wang2025moge}, and dynamic scene reconstructions \cite{zhang2024monst3r}, although not achieving real-time performance. 
To address these limitations, Spann3R \cite{wang20243d} introduces an incremental reconstruction approach to enable fast real-time operation. However, it rapidly accumulates drift errors due to the lack of global alignment and yields poor reconstruction accuracy. On the same track, SLAM3R \cite{liu2025slam3r} revisits SLAM by proposing a real-time 3D reconstruction method that does not  require explicit pose estimation. Nonetheless, current 3R-based frameworks remain focused on detailed geometry reconstruction, disregarding higher-level scene understanding. 

\begin{figure*}[t]
	\centering
	\includegraphics[width=0.9\linewidth,scale=1.00]{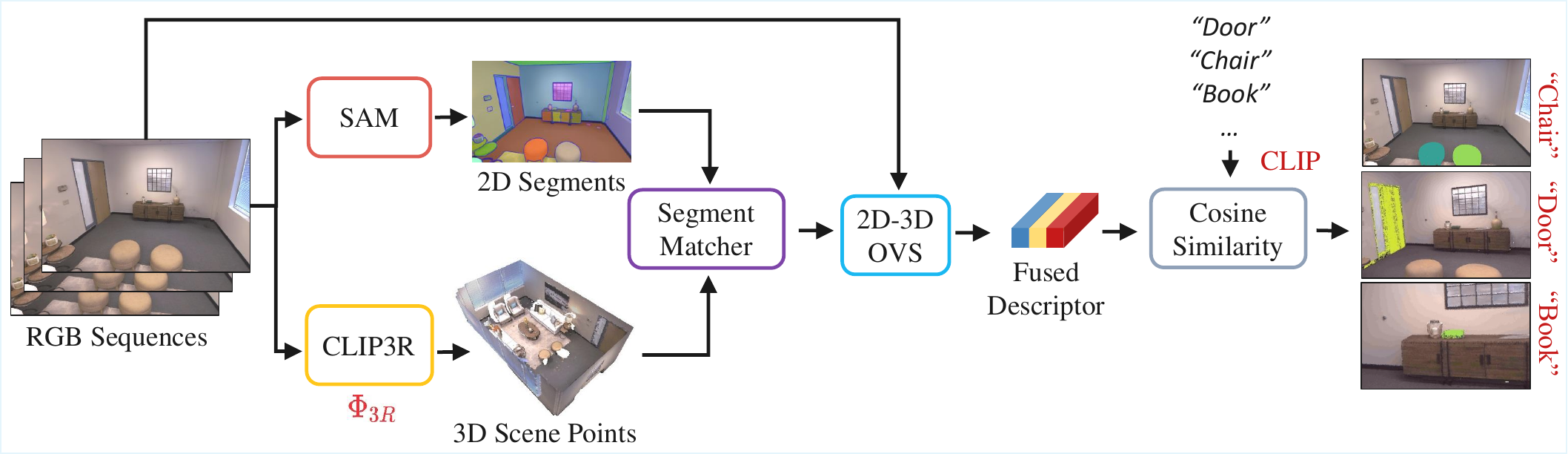}\vspace{-0.3cm}
	\caption{\textbf{Overview of Ov3R}. Given RGB-only videos, we first apply CLIP3R to produce scene points while SAM predicts 2D segments. Each 2D segment is matched to its corresponding 3D points to obtain 3D semantics. Next, the 2D-3D OVS extracts the fused 2D-3D descriptor to compute the cosine similarity with the text embeddings corresponding to a set of semantic classes.} 

	\label{fig:overview}
\end{figure*}

\subsection{3D Semantics}


\textbf{Closed-Vocabulary Online Semantics}. These methods predict per-pixel segmentation using pre-defined classes. SemanticFusion \cite{mccormac2017semanticfusion} pioneered per-pixel 3D semantic prediction from multi-view images. Recently, several semantic SLAM methods have been introduced to simultaneously reconstruct dense maps and semantics. NIS-SLAM \cite{zhai2024nis} integrates semantics into tri-plane features to optimize both scene representation and semantic predictions. SGS-SLAM \cite{li2024sgs} and NEDS-SLAM \cite{ji2024neds} encode semantics into Gaussians and incrementally reconstruct scenes with a closed-set segmentation. These methods achieve scene understanding by assuming a known set of 2D semantic annotations. However, relying on pre-defined semantic classes limits their applicability in real-world scenarios where novel object categories may appear.

\textbf{3D Open-Vocabulary Offline Semantics}. Most open-vocabulary semantic methods rely heavily on pre-processed 3D scene points as input. OpenScene \cite{peng2023openscene} encodes per-pixel dense features in CLIP feature space and associates 2D dense features with known 3D scene points. Open3DIS \cite{nguyen2024open3dis} relies significantly on pre-processed 2D and 3D semantic instances to extract pointwise features. Unlike these approaches that require known 3D points, OpenNeRF \cite{engelmann2024opennerf} directly uses RGB images as input for NeRF and OpenSeg \cite{ghiasi2022scaling}.
Similarly, HOV-SG \cite{werby2024hierarchical} extracts 3D semantics and scene graph hierarchy from RGB-D sequences, relying heavily on feature pre-processing. Unlike the above offline methods, OVO \cite{martins2024ovo} proposes the first online 3D open-vocabulary semantic mapping compatible with SLAM systems. It can extract and track 3D segments from RGB-D videos while encoding CLIP descriptors through a proposed CLIP merging strategy. However, it still requires depth information to be collected by an active sensor and explicit pose estimation from a SLAM system running concurrently. 
Besides, OVO computes CLIP descriptors solely from 2D images, overlooking the potential of 3D geometric information. 
In contrast, our Ov3R can extract 3D segments from RGB-only videos, not requiring explicit camera tracking. 

\section{Method}

We now introduce Ov3R, whose overall architecture is depicted in Figure \ref{fig:overview}. 
It consists of two main components, highlighted in yellow and blue: (i) a CLIP-informed 3R-based model (CLIP3R) and (ii) a 2D-3D OVS module. {CLIP3R produces dense scene points with semantic alignment, while 2D-3D OVS processes the semantic features predicted by the former for the downstream task of open-vocabulary 3D semantic segmentation. It enables Ov3R to reconstruct 3D semantic scenes while retaining the possibility to perform two tasks either jointly or independently. }


\subsection{CLIP3R}
\begin{figure*}[t]
	\centering
	\includegraphics[width=0.9\linewidth]{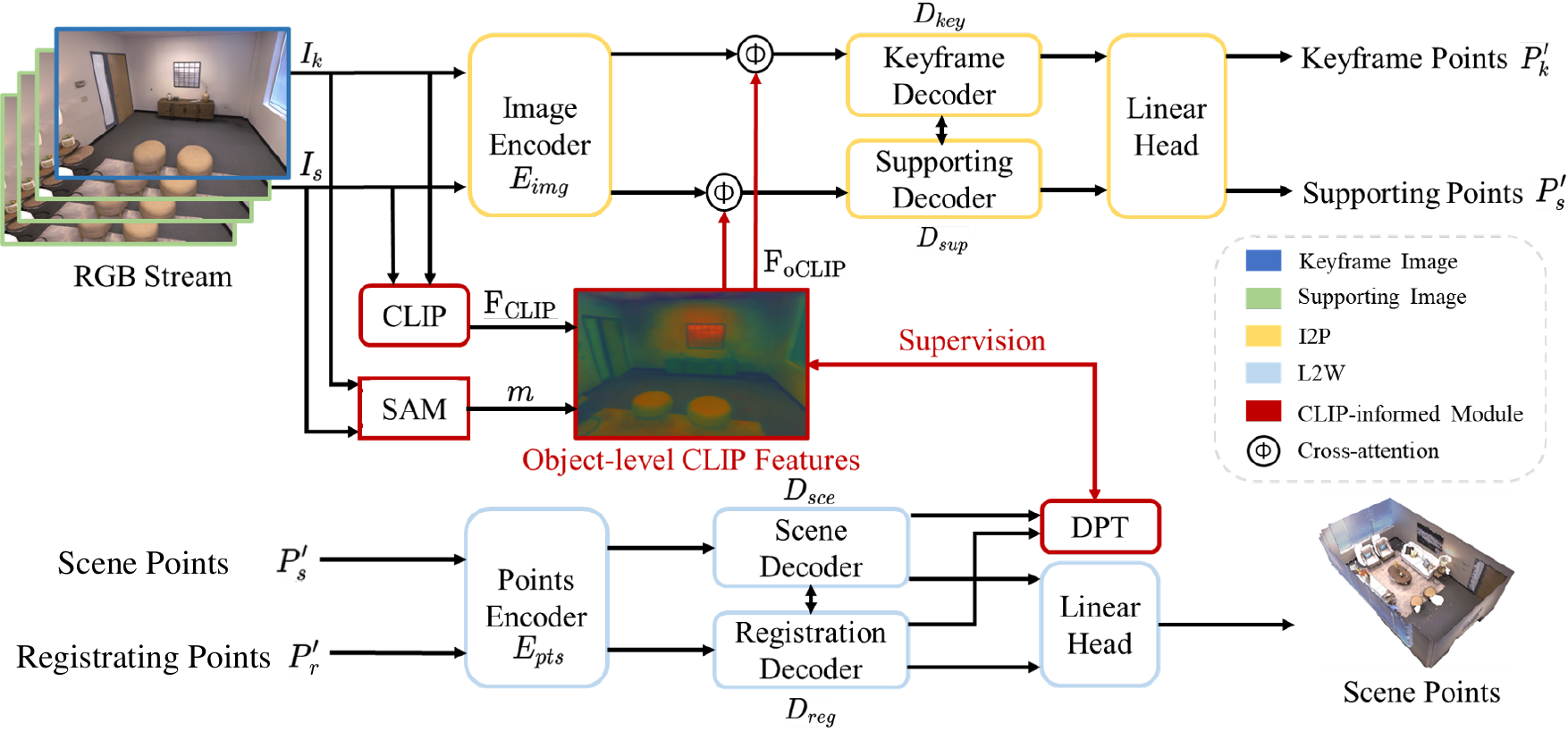}\vspace{-0.3cm}
	\caption{\textbf{CLIP3R Overview.} I2P integrates object-level CLIP features with visual embeddings to predict local pointmaps. L2W then aligns these local pointmaps to global scene coordinates while predicting object-level CLIP3R features with a DPT head in scene space.}

	\label{fig:clip3r}
\end{figure*}

We introduce CLIP3R, a CLIP-informed 3D reconstruction model that integrates the rich semantic understanding embedded in CLIP features and enables open-vocabulary semantic segmentation as a downstream task thanks to the image-text alignment. 
It can be formalized as a model $\Phi _{3R}$ that processes multi-view RGB sequences $I_{i}^{(H\times W \times 3)}$ to predict dense point maps $P_{i}^{(H\times W \times 3)}$:
\begin{equation}
    \Phi _{3R}\left ( I_{i}^{(N\times H\times W \times 3)}   \right )\to P_{i}^{(H\times W \times 3)} , i=1,..., L  
\end{equation}
As shown in Figure \ref{fig:clip3r}, inspired by \cite{liu2025slam3r}, CLIP3R deploys two main branches: an Image-to-Points (I2P, in yellow) model and a Local-to-World (L2W, in green) model. 

1) The former is a ViT, inspired by DUSt3R, running over a local window of images to predict a pointmap aligned to the reference system of the central frame (i.e., the keyframe). This is achieved by processing images with a shared encoder $E_{img}$ and two decoders $D_{key}$ and $D_{sup}$ for the keyframe and remaining images, respectively. Notably, no explicit camera pose estimation is performed, which in turn can be derived from the pointmap as a byproduct.

2) The latter, instead, processes the local pointmaps and aligns them to the scene-level reconstruction. It employs a pointmap encoder $E_{pts}$, a registration decoder $D_{reg}$, and a scene decoder $D_{sce}$, sharing the same structure as the encoder and decoder used by I2P. It identifies correlated keyframes using a reservoir and retrieval strategy \cite{liu2025slam3r} and processes them to predict global scene points $\hat{P'_{i}}$.

Both branches, however, lack high-level reasoning about scene semantics, which we address through integration of CLIP features and CLIP-semantic supervision, to learn stronger semantic-geometric relationships and consistency.


\textbf{Object-level CLIP3R features}. 
Although rich in semantics, CLIP embeddings are image-level features and therefore fail at modeling the fine-grained semantics of individual objects in the scene \cite{zuo2025fmgs}. 
To overcome this limitation, we extract object-level CLIP3R features $\mathbf{F}_{\text{oCLIP}}^{\left ( H \times W \times D \right ) }$ in place of vanilla CLIP features. This is achieved by first obtaining $M$ object masks $m^{\left ( H \times W \times 1 \right ) }$ using the Segment Anything Model (SAM) \cite{kirillov2023segment}. We then create $M$ masked images and process them to extract CLIP patch embeddings, which are averaged and upsampled to $\mathbf{F}_{\text{CLIP}}^{\left ( H \times W \times D \right ) }$ to match the image resolution. Finally, the object-level features $\mathbf{F}_{\text{oCLIP}}^{\left ( H \times W \times D \right ) }$ are obtained by combining the individual CLIP features obtained from the $M$ masked images within a single features map as:

\begin{equation}
    \mathbf{F}_{\text{oCLIP}}^{\left ( H \times W \times D \right ) } = \sum_{i=0}^M m_i^{\left ( H \times W \times 1 \right )} \cdot \mathbf{F}_{\text{CLIP}_i}^{\left ( H \times W \times D \right )}
\end{equation}
These finer-level features are then integrated into CLIP3R, both in I2P and L2W branches in different ways.


\begin{figure*}[t]
	\centering
	\includegraphics[width=0.9\linewidth,scale=1.00]{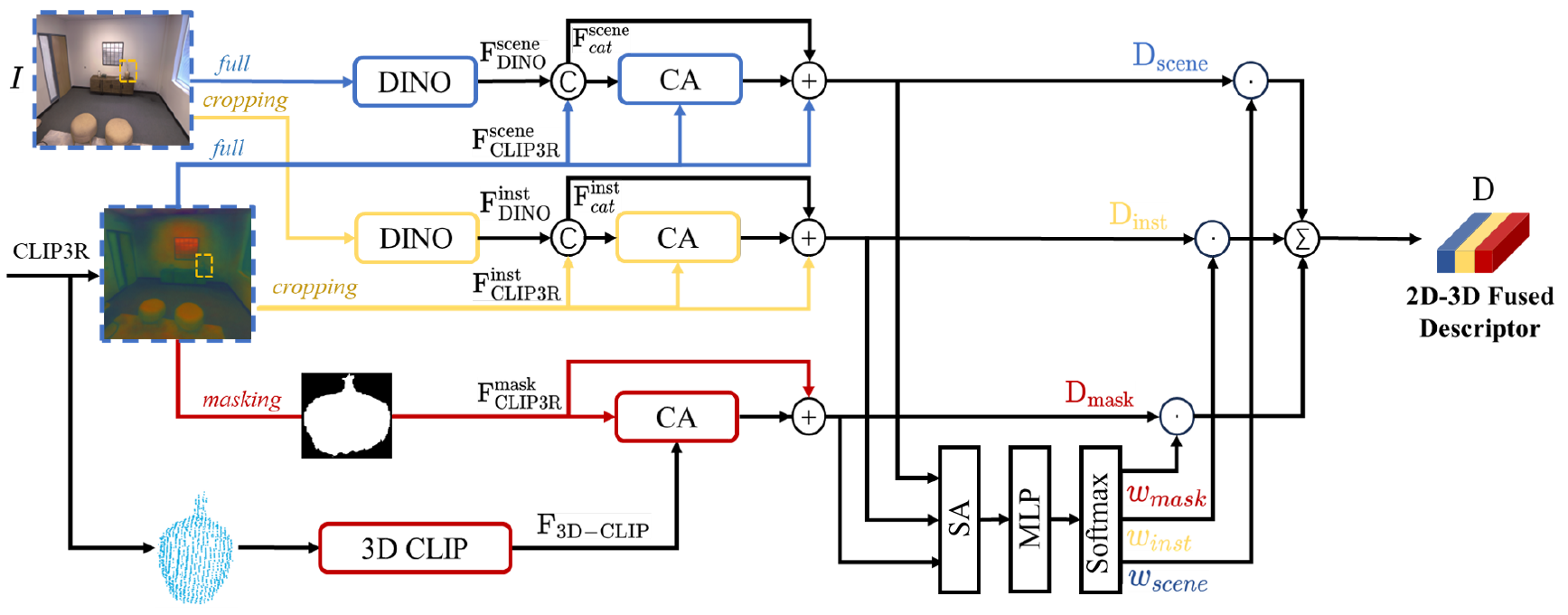}\vspace{-0.3cm}
	\caption{\textbf{2D-3D OVS Overview.} After matching 2D and 3D segments across images and pointmaps, CLIP3R, DINO, and 3D-CLIP features are combined into a 2D-3D fused descriptor, on top of which open-vocabulary semantic segmentation is performed. CLIP3R and DINO features are processed both at scene and instance levels. Meanwhile, 3D-CLIP features are extracted from masked 3D object points. } 
	\label{fig:fuseov}\vspace{-0.2cm}
\end{figure*}

\textbf{CLIP-informed I2P}. To predict local pointmaps, the I2P branch extracts visual tokens $\mathbf{F}_\text{ViT}^{\left ( T\times d \right )}$ from each frame in the processed window through ViT blocks \cite{dosovitskiy2020image}. To enrich I2P with knowledge from CLIP, we concurrently extract object-level CLIP3R features and tokenize them into $\mathbf{F}_{\text{oCLIP}} ^{\left ( T\times d \right )}$. 
Then, $\mathbf{F}_{\text{oCLIP}}$ and $\mathbf{F}_{vit}$ are processed through cross-attention to obtain the fused features $\mathbf{F}_\text{fuse}$, which are added to $\mathbf{F}_\text{ViT}$. The encoding process is denoted as:
\begin{equation}
\mathbf{F}_\text{fuse}=\mathbf{F}_\text{ViT}+\texttt{softmax}(\frac{\mathbf{F}_\text{ViT}\cdot\mathbf{F}_\text{oCLIP}^{T}}{\sqrt{d } } )\cdot \mathbf{F}_\text{oCLIP}
\end{equation}
where $d$ denotes the feature dimension. These features are then processed by the keyframe decoder $D_{key}$ and the supporting decoder $D_{sup}$ from the original I2P. A linear head is applied to predict both scene points $P'_{i}$ ($P'_{k}, P'_{s}\in P'_{i}$) and confidence map $C$. The revised I2P network is trained end-to-end through a confidence-aware loss over ground truth scene points: 
\begin{equation}
    \mathcal{L}_\text{I2P}=\sum_{i=1}^{L}M_{i}\cdot (C\cdot ||\frac{1}{z'}P'_{i} - \frac{1}{z}P_{i}||_1 -\alpha \log C)   
\end{equation}
where $L$ is the number of supporting frames, $M_{i}$ are the points with ground truth values in $P_{i}$, $z$ and $z'$ are scale factors, and $\alpha$ controls the confidence-based regularization term.

\textbf{CLIP-informed L2W}. The L2W branch maintains its original structure but incorporates an additional prediction head for object-level CLIP3R features, we refer to as $\mathbf{F}_\text{CLIP3R}$. This head embeds fine-grained semantic knowledge directly into the final 3D reconstruction while enforcing semantic consistency across the scene. This head is implemented as a DPT \cite{ranftl2021vision}, and the entire L2W branch is trained using two losses: $\mathcal{L}_{L2W}$ and $\mathcal{L}_{oCLIP}$. The former is similar to the loss used to supervise I2P:


\begin{equation}
    \mathcal{L}_\text{L2W}=\sum_{i=1}^{L}M_{i}\cdot (\hat{C}\cdot || \hat{P'_{i}} - P_{i}  ||_1 -\alpha \log\hat{C})   
\end{equation}
while the latter minimizes the feature distance between extracted $\mathbf{F}_\text{oCLIP}$ and predicted $\mathbf{F}_\text{CLIP3R}$ features:
\begin{equation}
    \mathcal{L}_\text{oCLIP}=\left \| \mathbf{F}_\text{CLIP3R} - \mathbf{F}_\text{oCLIP} \right \|_{1}  
\end{equation}

\subsection{2D-3D OVS}

We now introduce the second core component of Ov3R, responsible for performing open-vocabulary 3D semantic segmentation over the 3D reconstruction produced by CLIP3R.

\textbf{3D Segments Matching}. We revise and improve the strategy from \cite{martins2024ovo}, replacing CLIP features with those predicted by CLIP3R: after predicting dense point maps for each frame, we project the 3D point map predicted by CLIP3R onto each keyframe to assign 2D segment labels to the matched 3D points, corresponding to the SAM-predicted masks already used by CLIP3R itself. After processing all 2D segments, 3D points sharing the same label are merged into coherent 3D segments. 
Although this strategy effectively selects individual semantic components in the scene, it relies solely on 2D information and lacks explicit semantic knowledge from the 3D geometry itself. To address this limitation, we introduce 2D-3D fused descriptors, obtained as follows.

\textbf{2D-3D Fused Descriptors}. 
CLIP3R features alone are insufficient to model semantics in 3D space, as they 
are learned from 2D CLIP features and thus lack fine-grained 3D structural information. On the same track, while DINO \cite{caron2021emerging} features extract clear boundaries between inter-components, these lack knowledge about 3D geometry -- which is, on the contrary, available from point features \cite{hegde2023clip} extracted from 3D pointclouds. Therefore, we introduce a 2D-3D fused descriptor that combines these three complementary feature types extracted from i) CLIP3R, ii) DINO, and iii) a 3D-CLIP encoder \cite{hegde2023clip}: 

i) To capture the relationship between local and global cues, we adopt the multi-level architecture by \cite{werby2024hierarchical} and \cite{martins2024ovo}, as shown in Figure \ref{fig:fuseov}. From each keyframe, we leverage features predicted by CLIP3R (middle branch in the figure) at both scene-level (blue arrows in figure) and object instance-level (yellow arrows in figure), the latter by cropping the former according to 2D masks by SAM -- referred to as $\mathbf{F}_\text{CLIP3R}^\text{scene}$ and $\mathbf{F}_\text{CLIP3R}^\text{inst}$;

ii) In parallel, DINO features are extracted (top branch in the figure) and again processed at both the scene and instance levels, similarly to CLIP3R features. These are referred to, respectively, as $\mathbf{F}_\text{DINO}^\text{scene}$ and $\mathbf{F}_\text{DINO}^\text{inst}$;

iii) Finally, instance-level 3D features are extracted from the point map after retaining only 3D points associated with the same object label (bottom branch in the figure). A 2D counterpart of the object-level features is retrieved from CLIP3R, by masking the cropped instance-level features according to 2D segmentation masks corresponding to the same object label (red arrow).

\begin{table*}[t]
\centering
\resizebox{\textwidth}{!}{%
\begin{tabular}{lccccccccccr}
\toprule
Method &
  \begin{tabular}[c]{@{}c@{}}Room0\\ Acc. / Comp.\end{tabular} &
  \begin{tabular}[c]{@{}c@{}}Room1\\ Acc. / Comp.\end{tabular} &
  \begin{tabular}[c]{@{}c@{}}Room2\\ Acc. / Comp.\end{tabular} &
  \begin{tabular}[c]{@{}c@{}}Office0\\ Acc. / Comp.\end{tabular} &
  \begin{tabular}[c]{@{}c@{}}Office1\\ Acc. / Comp.\end{tabular} &
  \begin{tabular}[c]{@{}c@{}}Office2\\ Acc. / Comp.\end{tabular} &
  \begin{tabular}[c]{@{}c@{}}Office3\\ Acc. / Comp.\end{tabular} &
  \begin{tabular}[c]{@{}c@{}}Office4\\ Acc. / Comp.\end{tabular} &
  \begin{tabular}[c]{@{}c@{}}Average\\ Acc. / Comp.\end{tabular} &
  ATE RMSE &
  FPS \\ \midrule
DUSt3R &
  3.47 / \colorbox[HTML]{FFF3BB}{2.50} &
  \colorbox[HTML]{B7D3B7}{\textbf{2.53}} / \colorbox[HTML]{D8E8C5}{1.86} &
  \colorbox[HTML]{FFF3BB}{2.95} / \colorbox[HTML]{B7D3B7}{\textbf{1.76}} &
  4.92 / 3.51 &
  \colorbox[HTML]{FFF3BB}{3.09} / \colorbox[HTML]{B7D3B7}{\textbf{2.21}} &
  4.01 / 3.10 &
  \colorbox[HTML]{FFF3BB}{3.27} / \colorbox[HTML]{B7D3B7}{\textbf{2.25}} &
  \colorbox[HTML]{FFF3BB}{3.66} / \colorbox[HTML]{FFF3BB}{2.61} &
  \colorbox[HTML]{D8E8C5}{3.49} / \colorbox[HTML]{D8E8C5}{2.48} &
  4.76 &
  \textless{}1 \\
MASt3R &
  4.01 / 4.10 &
  3.61 / 3.25 &
  3.13 / 2.15 &
  \colorbox[HTML]{B7D3B7}{\textbf{2.57}} / \colorbox[HTML]{B7D3B7}{\textbf{1.63}} &
  12.85 / 8.13 &
  \colorbox[HTML]{B7D3B7}{\textbf{3.13}} / \colorbox[HTML]{B7D3B7}{\textbf{1.99}} &
  4.67 / 3.15 &
  3.69 / \colorbox[HTML]{D8E8C5}{2.47} &
  4.71 / 3.36 &
  1.67 &
  \textless{}\textless{}1 \\
\midrule
NICER-SLAM &
  \colorbox[HTML]{B7D3B7}{\textbf{2.53}} / 3.04 &
  3.93 / 4.10 &
  3.40 / 3.42 &
  5.49 / 6.09 &
  3.45 / 4.42 &
  4.02 / 4.29 &
  3.34 / 4.03 &
  \colorbox[HTML]{D8E8C5}{3.03} / 3.87 &
  3.65 / 4.16 &
  1.88 &
  \textless{}\textless{}1 \\
DROID-SLAM &
  12.18 / 8.96 &
  8.35 / 6.07 &
  3.26 / 16.01 &
  \colorbox[HTML]{D8E8C5}{3.01} / 16.19 &
  \colorbox[HTML]{B7D3B7}{\textbf{2.39}} / 16.20 &
  5.66 / 15.56 &
  4.49 / 9.73 &
  4.65 / 9.63 &
  5.50 / 12.29 &
  \colorbox[HTML]{D8E8C5}{0.33} &
  20 \\
DIM-SLAM &
  14.19 / 6.24 &
  9.56 / 6.45 &
  8.41 / 12.17 &
  10.16 / 5.95 &
  7.86 / 8.33 &
  16.50 / 8.28 &
  13.01 / 6.77 &
  13.08 / 8.62 &
  11.60 / 7.85 &
  0.46 &
  3 \\
  
MGS-SLAM &
  - &
  - &
  - &
  - &
  - &
  - &
  - &
  - &
  7.51 / 3.64 &
  \colorbox[HTML]{B7D3B7}{\textbf{0.32}} &
  0.6 \\
GO-SLAM &
  - &
  - &
  - &
  - &
  - &
  - &
  - &
  - &
  3.81 / 4.79 &
  \colorbox[HTML]{FFF3BB}{0.39} &
  8 \\
  \midrule
Spann3R &
  9.75 / 12.94 &
  15.51 / 12.94 &
  7.28 / 8.50 &
  5.46 / 18.75 &
  5.24 / 16.64 &
  9.33 / 11.80 &
  16.00 / 9.03 &
  13.97 / 16.02 &
  10.32 / 13.33 &
  32.79 &
  \textbf{\textgreater{}50} \\
SLAM3R &
  \colorbox[HTML]{FFF3BB}{3.19} / \colorbox[HTML]{D8E8C5}{2.40} &
  \colorbox[HTML]{FFF3BB}{3.12} / \colorbox[HTML]{FFF3BB}{2.34} &
  \colorbox[HTML]{D8E8C5}{2.72} / \colorbox[HTML]{FFF3BB}{2.00} &
  4.28 / 2.60 &
  3.17 / 2.34 &
  \colorbox[HTML]{FFF3BB}{3.84} / \colorbox[HTML]{FFF3BB}{2.78} &
  3.90 / 3.16 &
  4.32 / 3.36 &
  \colorbox[HTML]{FFF3BB}{3.57} / \colorbox[HTML]{FFF3BB}{2.62} &
  6.61 &
  24 \\
StreamVGGT &
  9.01 / 4.37 &
  12.22 / 4.66 &
  5.61 / 2.61 &
  6.64 / 3.45 &
  5.14 / 2.55 &
  12.84 / 7.23 &
  12.09 / 6.59 &
  15.48 / 6.35 &
  9.88 / 4.73 &
  12.50 &
  \textless{}1 \\
VGGT-SLAM &
  3.23 / 2.66 &
  18.92 / 15.41 &
  15.93 / 12.37 &
  4.01 / \colorbox[HTML]{FFF3BB}{2.49} &
  \colorbox[HTML]{D8E8C5}{3.01} / \colorbox[HTML]{FFF3BB}{2.32} &
  6.93 / 5.31 &
  \colorbox[HTML]{B7D3B7}{\textbf{2.97}} / \colorbox[HTML]{FFF3BB}{2.53} &
  5.19 / 3.83 &
  7.52 / 5.86 &
  7.10 &
  \textbf{\textgreater{}50} \\
\textbf{Ov3R (Ours)} &
  \colorbox[HTML]{D8E8C5}{2.67} / \colorbox[HTML]{B7D3B7}{\textbf{1.98}} &
  \colorbox[HTML]{D8E8C5}{2.96} / \colorbox[HTML]{B7D3B7}{\textbf{1.83}} &
  \colorbox[HTML]{B7D3B7}{\textbf{2.67}} / \colorbox[HTML]{D8E8C5}{1.91} &
  \colorbox[HTML]{FFF3BB}{3.41} / \colorbox[HTML]{D8E8C5}{2.24} &
  3.24 / \colorbox[HTML]{D8E8C5}{2.23} &
  \colorbox[HTML]{D8E8C5}{3.24} / \colorbox[HTML]{D8E8C5}{2.16} &
  \colorbox[HTML]{D8E8C5}{3.21} / \colorbox[HTML]{D8E8C5}{2.36} &
  \colorbox[HTML]{B7D3B7}{\textbf{2.99}} / \colorbox[HTML]{B7D3B7}{\textbf{2.25}} &
  \colorbox[HTML]{B7D3B7}{\textbf{3.05}} / \colorbox[HTML]{B7D3B7}{\textbf{2.12}} &
  6.00 &
  15 \\ \bottomrule
\end{tabular}%
 }\vspace{-0.3cm}
 \caption{\textbf{3D reconstruction and tracking results on Replica}. Methods are grouped into: 3R methods with low FPS, SLAM approaches, and real-time 3R models. Here, ``Ov3R" refers to the CLIP3R module. We highlight the \colorbox[HTML]{B7D3B7}{\textbf{first}}, \colorbox[HTML]{D8E8C5}{second}, and \colorbox[HTML]{FFF3BB}{third} best results.}\vspace{-0.3cm}
\label{tab:replica3r}
\end{table*}

\begin{figure*}[t]
	\centering
	\includegraphics[width=\linewidth,scale=1.00]{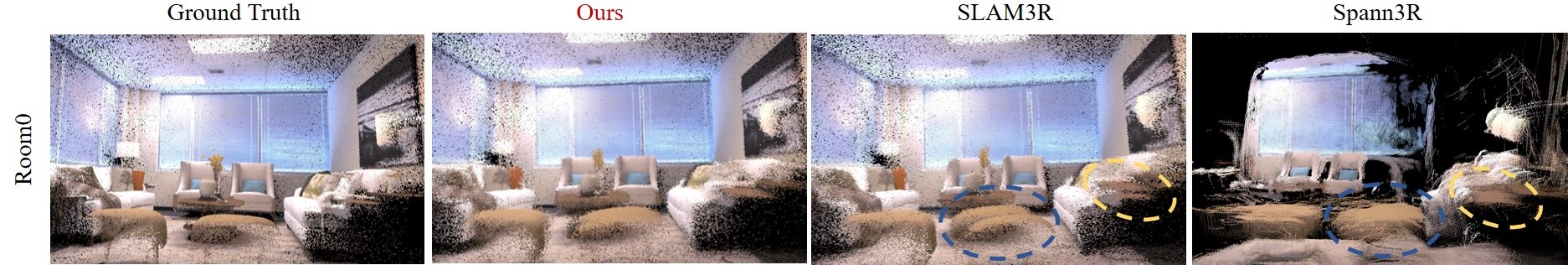}\vspace{-0.3cm}
	\caption{\textbf{Qualitative results -- Dense Pointmaps on Replica.} Compared to competing methods, Ov3R demonstrates superior completeness and geometric alignment, particularly visible in the reconstruction of \textbf{chairs (blue)} and \textbf{desks (yellow)}. 
 }\label{fig:recon}\vspace{-0.2cm}
\end{figure*}

Therefore, 2D-aware features by DINO and CLIP3R are concatenated at both the scene and the instance levels independently (\textcopyright{}  in the figure), after projecting the former to the same feature dimension as the latter by a linear layer. Concatenated features are further processed by a linear layer
\begin{equation}
\mathbf{F}_{cat}^\text{scene} = \texttt{Linear}([\texttt{Linear}(\mathbf{F}_\text{DINO}^\text{scene}), \mathbf{F}_\text{CLIP3R}^\text{scene}])
\end{equation}
\begin{equation}
\mathbf{F}_{cat}^\text{inst} = \texttt{Linear}([\texttt{Linear}(\mathbf{F}_\text{DINO}^\text{inst}), \mathbf{F}_\text{CLIP3R}^\text{inst}])
\end{equation}

\begin{table*}[t]
\centering
\resizebox{\textwidth}{!}{%
\begin{tabular}{lcccccccccr}
\toprule
Method &
  \begin{tabular}[c]{@{}c@{}}Chess\\ Acc. / Comp.\end{tabular} &
  \begin{tabular}[c]{@{}c@{}}Fire\\ Acc. / Comp.\end{tabular} &
  \begin{tabular}[c]{@{}c@{}}Heads\\ Acc. / Comp.\end{tabular} &
  \begin{tabular}[c]{@{}c@{}}Office\\ Acc. / Comp.\end{tabular} &
  \begin{tabular}[c]{@{}c@{}}Pumpkin\\ Acc. / Comp.\end{tabular} &
  \begin{tabular}[c]{@{}c@{}}RedKitchen\\ Acc. / Comp.\end{tabular} &
  \begin{tabular}[c]{@{}c@{}}Stairs\\ Acc. / Comp.\end{tabular} &
  \begin{tabular}[c]{@{}c@{}}Average\\ Acc. / Comp.\end{tabular} &
  ATE RMSE &
  FPS \\ \midrule
DUSt3R &
  2.26 / 2.13 &
  1.04 / 1.50 &
  \colorbox[HTML]{D8E8C5}{1.66} /  \colorbox[HTML]{B7D3B7}{\textbf{0.98}} &
  4.62 / 4.74 &
  \colorbox[HTML]{D8E8C5}{1.73} / 2.43 &
  \colorbox[HTML]{FFF3BB}{1.95} / 2.36 &
  3.37 / 10.75 &
  \colorbox[HTML]{FFF3BB}{2.19} / 3.24 &
  8.02 &
  \textless{}1 \\
MASt3R &
  \colorbox[HTML]{FFF3BB}{2.08} / 2.12 &
  1.54 / 1.43 &
   \colorbox[HTML]{B7D3B7}{\textbf{1.06}} / \colorbox[HTML]{FFF3BB}{1.04} &
  3.23 / \colorbox[HTML]{FFF3BB}{3.19} &
  5.68 / 3.07 &
  3.50 / 3.37 &
   \colorbox[HTML]{B7D3B7}{\textbf{2.36}} / 13.16 &
  3.04 / 3.90 &
  \colorbox[HTML]{B7D3B7}{\textbf{6.28}} &
  \textless{}\textless{}1 \\
  \midrule
Spann3R &
  2.23 / \colorbox[HTML]{FFF3BB}{1.68} &
  \colorbox[HTML]{FFF3BB}{0.88} / \colorbox[HTML]{FFF3BB}{0.92} &
  2.67 /  \colorbox[HTML]{B7D3B7}{\textbf{0.98}} &
  5.86 / 3.54 &
  2.55 /  \colorbox[HTML]{B7D3B7}{\textbf{1.85}} &
  2.68 /  \colorbox[HTML]{B7D3B7}{\textbf{1.80}} &
  5.65 /  \colorbox[HTML]{FFF3BB}{5.15} &
  3.42 / \colorbox[HTML]{FFF3BB}{2.41} &
  11.70 &
  \textbf{\textgreater{}50} \\
SLAM3R &
  \colorbox[HTML]{D8E8C5}{1.63} / \colorbox[HTML]{D8E8C5}{1.31} &
  \colorbox[HTML]{D8E8C5}{0.84}/ \colorbox[HTML]{D8E8C5}{0.83} &
  2.95 / 1.22 &
  \colorbox[HTML]{D8E8C5}{2.32} / \colorbox[HTML]{D8E8C5}{2.26} &
  1.81 / \colorbox[HTML]{FFF3BB}{2.05} &
  \colorbox[HTML]{D8E8C5}{1.84} / \colorbox[HTML]{FFF3BB}{1.94} &
  4.19 / 6.91 &
  \colorbox[HTML]{D8E8C5}{2.13} / \colorbox[HTML]{D8E8C5}{2.34} &
  {8.41} &
  25 \\
StreamVGGT &
  2.33 / 3.32 &
  1.21 / 1.96 &
  2.61 /  2.56 &
  3.21 / 4.56 &
  1.90 /  2.36 &
  2.75 /  3.85 &
  \colorbox[HTML]{FFF3BB}{3.03} /  \colorbox[HTML]{D8E8C5}{2.88} &
  2.43 / 3.07 &
  8.25 &
  \textless{}1 \\
VGGT-SLAM &
  \colorbox[HTML]{FFF3BB}{2.08} / 3.71 &
  1.43 / 2.28 &
  \colorbox[HTML]{FFF3BB}{2.36} /  2.65 &
  \colorbox[HTML]{FFF3BB}{2.88} / 4.83 &
  \colorbox[HTML]{FFF3BB}{1.75} /  2.55 &
  2.77 /  4.01 &
  \colorbox[HTML]{D8E8C5}{2.56} /  \colorbox[HTML]{B7D3B7}{\textbf{2.69}} &
  2.26 / 3.25 &
  \colorbox[HTML]{D8E8C5}{6.58} &
  \textbf{\textgreater{}50} \\
\textbf{Ov3R (Ours)} &
   \colorbox[HTML]{B7D3B7}{\textbf{1.53}} /  \colorbox[HTML]{B7D3B7}{\textbf{1.26}} &
   \colorbox[HTML]{B7D3B7}{\textbf{0.79}} /  \colorbox[HTML]{B7D3B7}{\textbf{0.76}} &
  2.41 / \colorbox[HTML]{D8E8C5}{0.99} &
   \colorbox[HTML]{B7D3B7}{\textbf{1.95}} /  \colorbox[HTML]{B7D3B7}{\textbf{1.93}} &
   \colorbox[HTML]{B7D3B7}{\textbf{1.66}} / \colorbox[HTML]{D8E8C5}{1.89} &
   \colorbox[HTML]{B7D3B7}{\textbf{1.63}} / \colorbox[HTML]{D8E8C5}{1.82} &
  3.88 / 5.95 &
   \colorbox[HTML]{B7D3B7}{\textbf{1.98}} /  \colorbox[HTML]{B7D3B7}{\textbf{2.08}} &
  \colorbox[HTML]{FFF3BB}{7.71} &
  17 \\ \bottomrule
\end{tabular}%
}\vspace{-0.3cm}
\caption{\textbf{3D reconstruction and tracking results on 7Scenes}. The results are divided into two groups: 3R methods with low FPS, and real-time 3R models. Here, ``Ov3R" refers to the CLIP3R module.}\vspace{-0.3cm}
\label{tab:7scenes}
\end{table*}

\begin{table}[t]
\resizebox{\columnwidth}{!}{%
\begin{tabular}{lcccc}
\toprule
Method &
  \begin{tabular}[c]{@{}c@{}}All\\ mIoU / mAcc\end{tabular} &
  \begin{tabular}[c]{@{}c@{}}Head\\ mIoU / mAcc\end{tabular} &
  \begin{tabular}[c]{@{}c@{}}Common\\ mIoU / mAcc\end{tabular} &
  \begin{tabular}[c]{@{}c@{}}Tail\\ mIoU / mAcc\end{tabular} \\ \midrule
\multicolumn{5}{c}{\bf Geometry: GT} \\ \midrule
OpenScene   & 15.9 / 24.6 & 31.7 / 44.8 & 14.5 / 22.6 & 1.5 / 6.3   \\
OpenNeRF    & 20.4 / 31.7 & 35.4 / 46.2 & 20.1 / 31.3 & 5.8 / \colorbox[HTML]{FFF3BB}{17.6}  \\
HOV-SG      & 22.5 / 34.2 & 35.9 / 44.2 & \colorbox[HTML]{D8E8C5}{23.6} / \colorbox[HTML]{D8E8C5}{42.3} & \colorbox[HTML]{FFF3BB}{8.0} / 16.1  \\
Open3DIS    & \colorbox[HTML]{FFF3BB}{25.6} / \colorbox[HTML]{D8E8C5}{38.7} & \colorbox[HTML]{B7D3B7}{\textbf{49.7}} / \colorbox[HTML]{B7D3B7}{\textbf{64.4}} & \colorbox[HTML]{FFF3BB}{22.1} / \colorbox[HTML]{B7D3B7}{\textbf{42.4}} & 4.9 / 9.4   \\
O$_{2}$V-mapping & 7.8 / 13.9  & 19.3 / 27.5 & 4.0 / 6.7   & 0.1 / 7.5   \\
OVO-mapping & \colorbox[HTML]{D8E8C5}{26.5} / \colorbox[HTML]{FFF3BB}{35.8} & \colorbox[HTML]{FFF3BB}{41.0} / \colorbox[HTML]{FFF3BB}{52.3} & 21.1 / 31.8 & \colorbox[HTML]{D8E8C5}{17.6} / \colorbox[HTML]{D8E8C5}{23.2} \\
\textbf{Ov3R (Ours)}  & \colorbox[HTML]{B7D3B7}{\textbf{31.9}} / \colorbox[HTML]{B7D3B7}{\textbf{42.3}} & \colorbox[HTML]{D8E8C5}{44.3} / \colorbox[HTML]{D8E8C5}{56.6} & \colorbox[HTML]{B7D3B7}{\textbf{28.6}} / \colorbox[HTML]{FFF3BB}{39.0} & \colorbox[HTML]{B7D3B7}{\textbf{22.8}} / \colorbox[HTML]{B7D3B7}{\textbf{31.5}} \\ \midrule
\multicolumn{5}{c}{\bf Geometry: CLIP3R} \\ \midrule
OVO-CLIP3R & \colorbox[HTML]{D8E8C5}{25.6} / \colorbox[HTML]{D8E8C5}{34.9} & \colorbox[HTML]{D8E8C5}{39.3} / \colorbox[HTML]{D8E8C5}{51.0} & \colorbox[HTML]{D8E8C5}{20.5} / \colorbox[HTML]{D8E8C5}{31.1} & \colorbox[HTML]{D8E8C5}{17.1} / \colorbox[HTML]{D8E8C5}{22.8} \\
\textbf{Ov3R (Ours)}  & \colorbox[HTML]{B7D3B7}{\bf 30.4} / \colorbox[HTML]{B7D3B7}{\bf 41.2} & \colorbox[HTML]{B7D3B7}{\bf 42.8} / \colorbox[HTML]{B7D3B7}{\bf 54.9} & \colorbox[HTML]{B7D3B7}{\bf 26.8} / \colorbox[HTML]{B7D3B7}{\bf 38.0} & \colorbox[HTML]{B7D3B7}{\bf 21.7} / \colorbox[HTML]{B7D3B7}{\bf 30.8} \\ \bottomrule

\end{tabular}%
}\vspace{-0.3cm}
\caption{\textbf{Open-vocabulary 3D semantic segmentation results on Replica dataset}. 
On top: methods running on ground truth 3D reconstructions. At the bottom: methods running on CLIP3R reconstructions. Here, ``Ov3R" refers to the 2D-3D OVS module.
}\vspace{-0.3cm}
\label{tab:replicaov}
\end{table}
These features are cross-attended to original CLIP3R features at both levels independently (CA block in the figure). The resulting features are finally summed with the previous concatenated features and CLIP3R features ($\oplus$ in the figure).

\small
\begin{equation}
\mathbf{D}_\text{scene} = \mathbf{F}_\text{CLIP3R}^\text{scene} + \mathbf{F}_{cat}^\text{scene} + \texttt{softmax}(\frac{\mathbf{F}_\text{CLIP3R}^\text{scene}\cdot \mathbf{F}_{cat}^{\text{scene }T}}{\sqrt{d }} )\cdot\mathbf{F}_{cat}^\text{scene}
\end{equation}
\begin{equation}
\mathbf{D}_\text{inst} = \mathbf{F}_\text{CLIP3R}^\text{inst} + \mathbf{F}_{cat}^\text{inst} + \texttt{softmax}(\frac{\mathbf{F}_\text{CLIP3R}^\text{inst}\cdot \mathbf{F}_{cat}^{\text{inst }T}}{\sqrt{d }} )\cdot\mathbf{F}_{cat}^\text{inst}
\end{equation}
\normalsize In parallel, we introduce a CLIP-like distilled 3D point encoder \cite{hegde2023clip} to extract geometric features. This 3D encoder is pre-trained on triplets of point clouds, corresponding images, and text using natural language supervision. The 3D features obtained by this distilled encoder naturally align with the CLIP latent space. 
These 3D features are then processed by a linear layer and cross-attended with masked object-level CLIP3R features to learn semantic-geometric relationships, and finally summed to these latter again similarly to what done in the DINO branch. 
\small\begin{equation}
    \mathbf{D}_\text{mask} = \mathbf{F}_\text{CLIP3R}^\text{mask} + \texttt{softmax}(\frac{\textbf{F}_\text{CLIP3R}^\text{mask}\cdot \mathbf{F}_\text{3D-CLIP}^{T}}{\sqrt{d }} )\cdot \mathbf{F}_\text{3D-CLIP}
\end{equation}
\normalsize Finally, we employ the weighted merging strategy from \cite{martins2024ovo} to combine these three features from different levels:
\begin{equation}
    \mathbf{D} = 
    w_\text{scene}\odot \mathbf{D}_\text{scene} +
    w_\text{inst}\odot \mathbf{D}_\text{inst} +
    w_\text{mask}\odot \mathbf{D}_\text{mask}
\end{equation}
where $\mathbf{D}$ is our 2D-3D fused descriptor. $\odot$ denotes the Hadamard product. The weights $w_{i}$ are obtained by a shallow model comprising a self-attention layer, an MLP, and a softmax layer. We pre-trained our 2D-3D fusion model by minimizing the sigmoid cosine similarity loss:
\begin{equation}
    \mathcal{L}_\text{sim} = -\frac{1}{\left | B \right | } \sum_{i}^{\left | B \right | } \sum_{j}^{\left | B \right | }log\left ( \frac{1}{1+e^{z_{ij}(-kd_{i}\cdot t_{j} + b)}  }  \right )  
\end{equation}
where $B$ is a mini-batch containing pairs of 2D segments and semantic labels. $z_{ij}$ is a binary label (1 or -1) indicating whether the segment and semantic label are paired. $t_{j}$ is the text embedding encoded by CLIP. $k$ and $b$ are learnable bias and temperature parameters, respectively. 
{
At inference time, the similarity between fused descriptors and a set of text embeddings corresponding to semantic classes is computed to select the class with the highest similarity.
}
\section{Experiments}


\textbf{Datasets}. For the {3D reconstruction task}, we follow \cite{liu2025slam3r} and train CLIP3R on ScanNet++ \cite{yeshwanth2023scannet++}, Aria Synthetic Environments \cite{avetisyan2024scenescript}, and CO3D-v2 \cite{reizenstein2021common}, which provide diverse scenarios and objects from both real-world and synthetic scenes. We evaluate 3D reconstruction performance on Replica \cite{straub2019replica} and 7Scenes \cite{shotton2013scene}. For the {open-vocabulary 3D segmentation} task, we train the 2D-3D OVS module on ScanNet++ and evaluate it on Replica and ScanNetv2 \cite{dai2017scannet}, following \cite{martins2024ovo}. Additionally, we report camera tracking results from CLIP3R on both Replica and 7Scenes. 

\textbf{Metrics.} We adopt standard metrics including \emph{Accuracy (cm), completion (cm)} for 3D reconstruction, \emph{Absolute Trajectory Error (ATE RMSE)} for tracking accuracy, and \emph{Frame Per Second (FPS)} to assess efficiency. For open-vocabulary 3D segmentation, we follow OVO protocol \cite{martins2024ovo} and use \emph{mean Intersection Over Union (mIoU)} and \emph{mean Accuracy (mAcc)} between the ground truth 3D labels and the labeled vertices of 3D meshes. On ScanNetv2, we also evaluate label-frequency weighted metrics (f-mIoU and f-mAcc).

\textbf{Implementation Details.} Ov3R is trained on 4 NVIDIA A100 GPU (64GB), while inference can be performed on a single 3090 NVIDIA GPU (24GB). In CLIP3R, we set 24 encoder blocks and 12 decoder blocks with linear heads, setting the window length to $L=5$ for initialization and $L=11$ for subsequent incremental windows. The 2D-3D OVS model is trained for 15 epochs, with batch size 512. We use SAM 2.1 for 2D instance segmentation, SigLip ViT-SO400 for CLIP features, ViTS-16 for DINO features, and PointMLP-1024 for 3D point features.

\subsection{3D Reconstruction}

We start by assessing the quality of 3D reconstructions produced by Ov3R against state-of-the-art methods.

\textbf{Replica}. Table \ref{tab:replica3r} compares Ov3R against the state-of-the-art reconstruction methods on 8 scenes from Replica, including both SLAM-based and 3R-based approaches. Overall, Ov3R outperforms all state-of-the-art methods while maintaining up to 15 FPS processing speed.  Although Spann3R achieves up to 50 FPS, it fails to provide accurate and complete 3D reconstructions, as shown in Figure \ref{fig:recon}. In contrast, Ov3R achieves consistently superior reconstructions compared to both Spann3R and SLAM3R.

\textbf{7Scenes}. Table \ref{tab:7scenes} reports the evaluation on the 7Scenes dataset, carried out by uniformly sampling one-twentieth of the frames from each test sequence following \cite{liu2025slam3r}. As shown in the table, on average, Ov3R surpasses all of the baselines in both accuracy and completion, confirming the trends already highlighted on the Replica dataset. 

\begin{table}[t]
\resizebox{\columnwidth}{!}{%
\begin{tabular}{lcccccccc}
\toprule
\multirow{2}{*}{Methods} & \multicolumn{4}{c}{ScanNet20}               & \multicolumn{4}{c}{ScanNet200}                                \\ \cline{2-9} 
            & mIoU          & mAcc          & f-mIoU & f-mAcc & mIoU & mAcc & f-mIoU & f-mAcc \\ \midrule
\multicolumn{9}{c}{\bf Geometry: GT} \\ \midrule
OpenScene   & \colorbox[HTML]{B7D3B7}{\textbf{44.6}} & \colorbox[HTML]{B7D3B7}{\textbf{61.9}} & \colorbox[HTML]{D8E8C5}{57.6}   & \colorbox[HTML]{D8E8C5}{71.0}     & 9.4  & 12.6 & 27.8   & 32.0     \\
HOV-SG      & 34.4          & 51.1          & 47.3   & 61.8   & 11.2 & 18.7 & 27.7   & \colorbox[HTML]{FFF3BB}{37.6}   \\
Open3DIS    & 37.3          & \colorbox[HTML]{FFF3BB}{52.8}         & \colorbox[HTML]{FFF3BB}{57.0}    & 67.9   & \colorbox[HTML]{D8E8C5}{17.8} & \colorbox[HTML]{FFF3BB}{23.7} & \colorbox[HTML]{FFF3BB}{27.9}   & 34.1   \\
OVO-mapping & \colorbox[HTML]{FFF3BB}{38.1}          & 50.5          & \colorbox[HTML]{D8E8C5}{57.6}  & \colorbox[HTML]{FFF3BB}{70.5}   & \colorbox[HTML]{FFF3BB}{17.2} & \colorbox[HTML]{D8E8C5}{25.3} & \colorbox[HTML]{D8E8C5}{45.4}   & \colorbox[HTML]{D8E8C5}{56.4}   \\
\textbf{Ov3R(Ours)}               & \colorbox[HTML]{D8E8C5}{43.6} & \colorbox[HTML]{D8E8C5}{57.8} & \colorbox[HTML]{B7D3B7}{\textbf{65.0}} & \colorbox[HTML]{B7D3B7}{\textbf{79.4}} & \colorbox[HTML]{B7D3B7}{\textbf{19.4}} & \colorbox[HTML]{B7D3B7}{\textbf{28.5}} & \colorbox[HTML]{B7D3B7}{\textbf{50.8}} & \colorbox[HTML]{B7D3B7}{\textbf{61.6}} \\ \midrule
\multicolumn{9}{c}{\bf Geometry: CLIP3R} \\ \midrule
OVO-CLIP3R & \colorbox[HTML]{D8E8C5}{32.6}    & \colorbox[HTML]{D8E8C5}{44.2}     & \colorbox[HTML]{D8E8C5}{48.9}     & \colorbox[HTML]{D8E8C5}{64.8}   & \colorbox[HTML]{D8E8C5}{14.1}     & \colorbox[HTML]{D8E8C5}{21.5} & \colorbox[HTML]{D8E8C5}{38.0}   & \colorbox[HTML]{D8E8C5}{51.2}   \\
\textbf{Ov3R(Ours)}  & \colorbox[HTML]{B7D3B7}{\bf 37.0} & \colorbox[HTML]{B7D3B7}{\bf 48.1} & \colorbox[HTML]{B7D3B7}{\bf 56.2} & \colorbox[HTML]{B7D3B7}{\bf 68.1} & \colorbox[HTML]{B7D3B7}{\bf 17.5} & \colorbox[HTML]{B7D3B7}{\bf 25.8} & \colorbox[HTML]{B7D3B7}{\bf 45.6} & \colorbox[HTML]{B7D3B7}{\bf 56.3} \\ \bottomrule
\end{tabular}%
}\vspace{-0.3cm}
\caption{\textbf{Open-vocabulary 3D semantic segmentation results on ScanNetv2.} 
On top: methods running on ground truth 3D reconstructions. At the bottom: methods running on CLIP3R reconstructions. Here, ``Ov3R" refers to the 2D-3D OVS module.}\vspace{-0.3cm}
\label{tab:scannetov}
\end{table}

\begin{figure*}[htbp]
	\centering
	\includegraphics[width=\linewidth,scale=1.00]{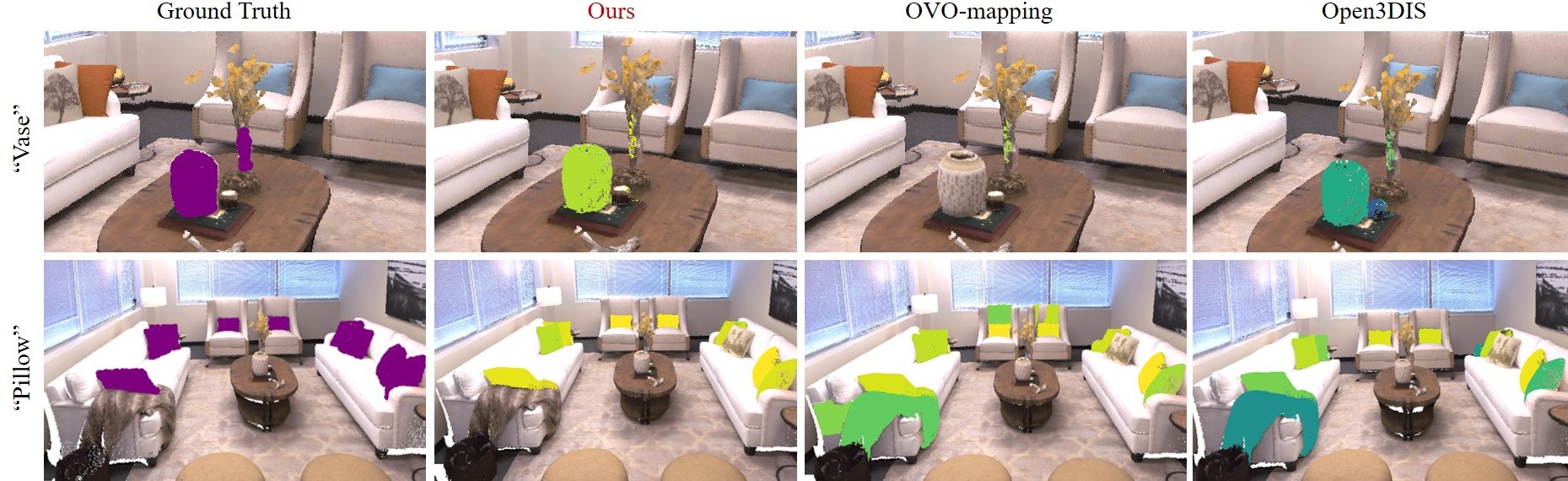}\vspace{-0.3cm}
	\caption{\textbf{Visualization of open-vocabulary 3D queries on the Replica}. We visualize two queries from \emph{"Head"} and \emph{"Common"} in \emph{room0}. The similarities of queries are shown from 
 \emph{low (blue)} to 
 \emph{mid (green)}, and 
 \emph{high (yellow)}. OVO-mapping yields wrong \emph{"Pillows"} and the missing \emph{"Vase"}, while Open3DIS also produces wrong \emph{"Pillows"} and low-similarity \emph{"Vase"}. }\vspace{-0.3cm}
	\label{fig:ov}
\end{figure*}

\subsection{Camera Tracking}

Although 3R models do not explicitly estimate trajectories, camera poses can be derived from their predicted scene points. Following \cite{wang2024dust3r}, we extract poses by running PnP-RANSAC \cite{fischler1981random} from the predicted 3D points, given the known camera intrinsics of each frame. 
Accordingly, we assess the tracking performance of Ov3R on the Replica and 7Scenes datasets. As shown in Table \ref{tab:replica3r} and Table \ref{tab:7scenes}, Ov3R outperforms the other two real-time 3R-based frameworks, Spann3R and SLAM3R, narrowing the gap between with the much slower dense SLAM systems. Only VGGT-SLAM achieves better results on 7Scenes, thanks to the SL(4)  optimization performed outside of the model. 

\subsection{Open-Vocabulary 3D Semantic Segmentation}

We continue our evaluation by focusing on the open-vocabulary 3D semantic segmentation task. 
We conduct experiments under two different settings: segmentation performed on geometry reconstructed from (i) ground truth depth maps, simulating an offline setting, and (ii) RGB-only videos processed by CLIP3R.

\begin{table}[t]
\renewcommand{\tabcolsep}{15pt}
\resizebox{\columnwidth}{!}{%
\begin{tabular}{lccc}
\toprule
                & Acc. & Comp. & RMSE \\ \midrule
(A) w/o CLIP-insert & 3.31 & 2.35  & 6.46 \\
(B) w/o CLIP head   & 3.20 & 2.21  & 6.32 \\
(C) w/ standard CLIP   & 3.18 & 2.20 & 6.28 \\
\bf Ov3R (all)      & \textbf{3.05} & \textbf{2.12}  & \textbf{6.00} \\ \bottomrule
\end{tabular}%
}\vspace{-0.3cm}
\caption{\textbf{Ablation study of CLIP3R on Replica}. We study the impact of CLIP-insertion in I2P (CLIP-insert) and the CLIP-semantic supervision in L2W (CLIP head). }\vspace{-0.3cm}
\label{tab:ablation3r}
\end{table}

\textbf{Replica}. We run experiments on the 8 standard scenes of Replica, annotated with 51 different semantic classes, and evaluate the results directly in 3D space rather than by projecting semantic masks onto 2D images \cite{kerr2023lerf}.
Following \cite{engelmann2024opennerf}, these labels are divided into three categories (\textit{"Head", "Common", "Tail"}) according to their frequency in the dataset. Thanks to its modular design, we compare Ov3R with both offline approaches and the recent online method, OVO \cite{martins2024ovo}. 
In Table \ref{tab:replicaov}, Ov3R outperforms all baselines on \emph{mIoU} and \emph{mAcc} across \emph{All} labels on average, achieving the best trade-off across all categories. Although Open3DIS is better on \emph{"Head"} labels, it fails on less frequent objects. 
As shown in Figure \ref{fig:ov}, Ov3R accurately identifies the queried classes, whereas OVO and Open3DIS yield inaccurate segmentation or miss objects entirely.


\textbf{ScanNetv2}. We continue the experiments on 5 scenes from ScanNetv2, using both the original (20) and extended (200) classes. Table \ref{tab:scannetov} shows that Ov3R surpasses all baselines on the ScanNet200 dataset, demonstrating its generalization capability even with diverse and previously unseen class annotations. For the ScanNet20 dataset, Ov3R outperforms other methods in terms of f-mIoU and f-mAcc over ground truth reconstructions, while achieving comparable results to OpenScene in mIoU and mAcc. Notably, OpenScene performance significantly deteriorates when evaluated on a much larger set of classes.
Consistently, Ov3R outperforms OVO in the online setting across both datasets.

\begin{table}[t]
\renewcommand{\tabcolsep}{10pt}
\resizebox{\columnwidth}{!}{%
\begin{tabular}{lcccc}
\toprule
               & mIOU           & mAcc           & f-mIoU         & f-mAcc         \\ \midrule
(D) w/o DINO       & 28.05          & 35.63          & 53.46          & 66.59          \\
(E) w/o 3D encoder & 28.46          & 36.20          & 52.81          & 65.38          \\
\bf Ov3R (all)     & \textbf{30.65} & \textbf{41.31} & \textbf{54.64} & \textbf{67.61} \\ \bottomrule
\end{tabular}%
}\vspace{-0.3cm}
\caption{\textbf{Ablation study of 2D-3D OV}. We report the advancement brought by different fusion strategies. }\vspace{-0.3cm}
\label{tab:ablationov}
\end{table}

\subsection{Ablation Study}

We conclude our experiments by assessing the impact of different sub-modules in both CLIP3R and 2D-3D OVS. 

\textbf{CLIP3R Component Analysis}. We first assess the effects of our CLIP-informed strategy. As shown in Table \ref{tab:ablation3r}, retaining only CLIP-semantic supervision (A) yields significant drops in both reconstruction accuracy and camera tracking, demonstrating the importance of CLIP cues in the reconstruction model. Moreover, removing the additional supervision (B) leads to consistently lower performance, confirming its key role in CLIP3R for 3D reconstruction. Finally, we demonstrate that using vanilla CLIP (C) in place of the object-level features also yields suboptimal results.

\textbf{2D-3D OVS Component Analysis}. 
Table \ref{tab:ablationov} shows that either CLIP descriptors fused with 3D features only (D) or those fused with DINO features only (E) lead to drops in accuracy. In contrast, our full 2D-3D OVS achieves the best performance, demonstrating that combining both 2D and 3D descriptors is crucial for accurate 3D segmentation. 

\subsection{Runtime Analysis}

Finally, we report a detailed runtime analysis concerning our entire framework in Table \ref{tab:efficiency}. Both our modules, CLIP3R and 2D-3D OVS, run at about 15 FPS, with the segmentation model represents the main bottleneck. However, we argue that replacing SAM2 with faster variants \cite{zhao2023fast} would allow Ov3R to meet real-time constraints.

\section{Conclusion}
We presented Ov3R, an open-vocabulary semantic 3D reconstruction framework. 
Compared to recent methods, Ov3R learns stronger semantic-geometric relationships and enforces semantic consistency through the proposed CLIP3R. Moreover, the introduced 2D-3D OVS enhances CLIP descriptors with spatial and geometric awareness. Our experiments demonstrate the superior accuracy and completeness of 3D reconstructions and 3D open-vocabulary semantic segmentation achieved by our approach. 

\textbf{Limitations and Future Work.} Ov3R inherits one of the limitations of 3R models, i.e., the suboptimal accuracy of the retrieved camera poses. Future research will aim to overcome this limitation by integrating techniques from the SLAM literature, such as global bundle adjustment.


{
    \small
    \bibliographystyle{ieeenat_fullname}
    \bibliography{main}
}


\end{document}